\documentclass[runningheads]{llncs}
\usepackage{subfigure}
\usepackage{graphicx}
\usepackage{graphics}
\usepackage{amsmath,amssymb} 
\usepackage{color}

\usepackage{hyphenat}

\begin{document}
	
	\newcommand{\point}{
		\raise0.7ex\hbox{.}
	}
	
	
	\pagestyle{headings}

\mainmatter

\title{Semi-Supervised Domain Adaptation for Weakly Labeled Semantic Video Object Segmentation} 
\titlerunning{Semi-Supervised Domain Adaptation for Weakly Labeled Semantic Video Object Segmentation}
\authorrunning{Huiling Wang, Tapani Raiko, Lasse Lensu, Tinghuai Wang, Juha Karhunen}

\author{Huiling Wang \inst{1}   Tapani Raiko \inst{1}   Lasse Lensu \inst{2}  Tinghuai Wang \inst{3} \and Juha Karhunen \inst{1}}
\institute{
	Aalto University, Finland
	\and
	Lappeenranta University of Technology, Finland
	\and
	Nokia Labs, Finland
}

\maketitle

\begin{abstract}
	
Deep convolutional neural networks (CNNs)  have been immensely successful in many high-level computer vision tasks 
given large labeled datasets.  However, for video semantic object segmentation, a domain where  labels are scarce, 
effectively exploiting the representation power of CNN with limited training data remains a challenge. Simply borrowing 
the existing pretrained CNN image recognition model for video segmentation task can severely hurt performance.  We propose a
semi-supervised  approach to adapting CNN image recognition model trained from labeled image data 
to the target domain exploiting both semantic evidence learned from CNN, and the intrinsic structures of video data.
By explicitly modeling and compensating for the domain shift from the source domain to the target 
domain, this proposed approach underpins a robust semantic object segmentation method 
against the changes in appearance, shape and occlusion in natural videos. We present extensive experiments on challenging 
datasets that demonstrate the superior performance of our approach compared with the state-of-the-art methods.

\end{abstract}

\section{Introduction}

\begin{figure}[t!]
\centering
\includegraphics[width=1.0\linewidth]{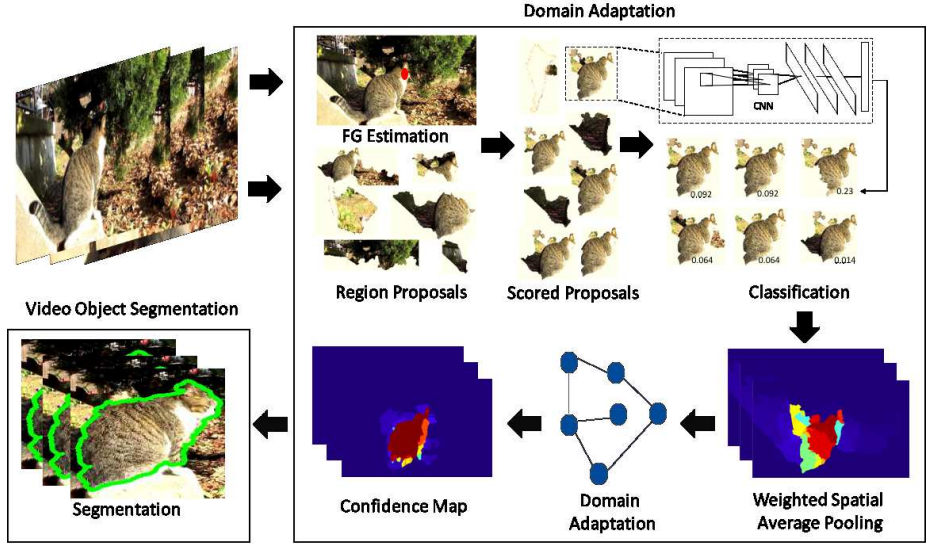}
\caption{Overview of our proposed method.}\label{fig:diagram}
\end{figure}

Semantically assigning each pixel in video with a known class label can be challenging for machines due to several reasons. Firstly, acquiring 
the prior knowledge about object appearance, shape or position is difficult.  Secondly,  gaining pixel-level annotation for training supervised 
learning algorithms is prohibitively expensive comparing with image-level labelling. Thirdly, background clutters, occlusion and object appearance 
variations  introduce visual ambiguities that in turn induce instability 
in boundaries and the potential for localised under- or over-segmentation. Recent years have seen encouraging progress, 
particularly in terms of generic object segmentation \cite{BroxM10,LeeKG11,ZhangJS13,Papazoglou2013,WangW14,wang2016primary}, and the success of convolutional neural networks in image recognition \cite{alexnet,vggnet,laddernet} also sheds light on semantic video object segmentation. 

Generic object segmentation methods \cite{LeeKG11,ZhangJS13,WangW14,Yang2015} largely utilise category independent region proposal methods 
\cite{EndresH10,ManenGG13}, to capture object-level description of the generic object in the scene incorporating motion 
cues. These approaches address the challenge of visual ambiguities to some extent, seeking the weak prior knowledge of what the object 
may look like and where it might be located. However, there are generally two major issues with these approaches. Firstly, the generic 
detection has very limited capability to determine the presence of an object. Secondly, such approaches are generally unable 
to determine and differentiate unique multiple objects, regardless of categories. These two bottlenecks limit these approaches to segmenting 
one single object or all foreground objects regardless classes or identifies.

Deep convolutional neural networks (CNNs) have been proven successful \cite{alexnet,vggnet,laddernet} in many high-level computer vision tasks such as image 
recognition and object detection. However, stretching this success to the domain of pixel-level classification or labelling, i.e., semantic segmentation,
is not naturally straightforward. This is not only owing to the difficulties of collecting pixel-level annotations, but also due to the nature
of large receptive fields of convolutional neural {\nobreak networks}. Furthermore, the aforementioned challenges present in video data demand
a data-driven representation of the video object in order to give a spatio-temporal coherent segmentation. This motivates us to develop a  framework for adapting image recognition models (e.g., CNN) trained on static images to a video domain for the demanding task of 
pixel labelling. This goal is achieved by proposing a semi-supervised domain adaptation approach to forming a data-driven object representation which  incorporates both the semantic evidence from pretrained CNN image recognition model and the constraint imposed by the intrinsic structure of video data. We exploit the constraint in video data that when the same object is recurring between video frames, the spatio-temporal coherence implies the associated unlabelled data to be the same label. This data-driven object representation underpins a robust object segmentation method for weakly labelled natural videos. 

The paper is structured as follows: We firstly review related work in video object segmentation (Sec. \ref{sec:literature}). Our proposed method is presented in
Sec.  \ref{sec:approach}, which is comprised of  domain adaptation and segmentation. Evaluations and comparisons in Sec. \ref{sec:evaluation} show the benefits of our method.  We conclude this paper with our findings in Sec. \ref{sec:conclusion}.

\section{Related Work}
\label{sec:literature}

Video object segmentation has received considerable attention in recent years, with the majority of research effort categorised into three  groups based on the level of supervisions: (semi-)supervised, unsupervised and weakly supervised methods. 

Methods in the first category normally require an initial annotation of the first frame, 
which either perform spatio-temporal grouping \cite{WangXSC04,CollomosseRH05} or propagate 
the annotation to drive the segmentation in successive frames \cite{WangC12,TsaiFNR12,Li2013,WangHC14}.

Unsupervised methods have been proposed as a consequence of the prohibitive cost of human-in-the-loop operations when processing ever-growing large-scale video data. Bottom-up approaches \cite{GrundmannKHE10a,XuXC12,Papazoglou2013} largely utilise spatio-temporal appearance and motion constraints, while motion segmentation approaches \cite{WangGP09,SundbergBMAM11} perform long-term motion analysis to cluster pixels or regions in video
data. Giordano {\em et al.} \cite{GiordanoMPS15} extended  \cite{Papazoglou2013} by introducing 
`perceptual organization' to improve segmentation. Taylor {\em et al.} \cite{TaylorKS15} inferred object segmentation through long-term occlusion relations, and introduced a numerical scheme to perform partition directly on pixel grid. Wang  {\em et al.} \cite{WangSP15} exploited saliency measure using geodesic distance to build global appearance models. Several methods \cite{LeeKG11,ZhangJS13,WangW14,Yang2015,wang2016primary} propose to introduce a top-down notion of object by exploring recurring object-like regions from still images by measuring generic object appearance (e.g., \cite{EndresH10}) to achieve state-of-the-art results. However, due to the limited recognition capability of generic object detection, these methods normally can only segment foreground objects regardless of semantic label.

The proliferation of user-uploaded videos which are frequently associated with semantic tags provides a vast resource for computer vision research. These semantic tags, albeit not spatially or temporally located in the video, suggest visual concepts appearing in the video. This social trend has led to an increasing interest in exploring the idea of segmenting video objects with weak supervision or labels. Hartmann {\em et al.} \cite{HartmannGHTKMVERS12} firstly formulated the problem as learning weakly supervised classifiers for a set of independent spatio-temporal segments.
Tang {\em et al.} \cite{TangSY013} learned discriminative model by leveraging labelled positive videos and a large collection of negative examples based on distance matrix. Liu {\em et al.} \cite{LiuTSRCB14} extended the traditional binary classification problem to multi-class and proposed nearest-neighbor-based label transfer algorithm which encourages smoothness between regions that are spatio-temporally adjacent and similar in appearance. Zhang {\em et al.} \cite{ZhangCLWX15} utilised pre-trained object detector to generate a set of detections and then pruned noisy detections and regions by preserving spatio-temporal constraints.

\section{Approach}
\label{sec:approach}
As shown in Fig. \ref{fig:diagram}, our method consists of two major components: domain adaptation and segmentation. Technical details of each component are provided in the following subsections.

\subsection{Domain Adaptation}

We set out our approach to first semantically discovering possible objects of interest from video. We then adapt the source domain from image recognition to  the target domain, i.e., pixel or superpixel level labelling.  This approach is built by additionally incorporating constraints obtained from a given similarity graph defined on unlabeled target instances.

\subsubsection{Proposal Scoring}

Unlike image classification or object detection, semantic object segmentation requires not only localising objects of interest within an image, but also assigning class label for pixels belonging to the objects. One potential challenge of using image classifier to detect objects is that any regions containing the object or even part of the object, might be ``correctly'' recognised, which results in a large search space to accurately localise the object. To narrow down 
the search of targeted objects, we adopt category-independent bottom-up object proposals.

As we are interested in producing segmentations and not just bounding boxes, we require region proposals. We consider those regions as candidate object hypotheses. The  objectness score associated with each proposal from \cite{EndresH10} indicates how likely it is for an image region contain an object of any class. However, this objectness score 
does {\em not} consider context cues, e.g. motion, object categories and temporal coherence etc., and reflects only the generic object-like properties of the region (saliency, apparent separation from background, etc.). We incorporate motion information as a context cue for video objects. There has been many previous works on estimating local motion cues and we adopt a motion boundary based approach as introduced in \cite{Papazoglou2013} which roughly produces a binary map indicating whether each pixel is inside the motion boundary after compensating camera motion. After acquiring the motion cues, we score each proposal $r$ by both appearance and context,  
\begin{equation} 
s_{r} = \mathcal{A}(r)+ \mathcal{C}(r)  \label{eq:objness1} \nonumber
\end{equation}
where $\mathcal{A}(r)$ indicates region level appearance score computed using \cite{EndresH10} and $\mathcal{C}(r)$ represents the contextual score of region $r$ which is defined as:
\begin{equation} 
\mathcal{C}(r) = \mathrm{Avg}(M^t(r))\cdot \mathrm{Sum}(M^t(r))  \label{eq:objness2} \nonumber
\end{equation}
where $\mathrm{Avg}(M^t(r))$ and $\mathrm{Sum}(M^t(r))$ compute the average and total amount of motion cues \cite{Papazoglou2013} included by proposal $r$ on frame $t$ respectively. Note that appearance, contextual and combined scores are normalised. 

\subsubsection{Proposal Classification}
\label{sec:classification1}
On each frame $t$ we have a collection of region proposals scored by their appearance and contextual information. These region proposals may contain various objects present in the video. In order to identify the objects of interest specified by the video level tag, region level classification is performed. We consider proven classification architectures such as VGG-16 nets \cite{vggnet} which did exceptionally well in ILSVRC14. VGG-16 net uses $3\times3$ convolution interleaved with max pooling and 3 fully-connected layers. 

In order to classify each region proposal, we firstly warp the image data in each region into a form that is compatible with the CNN (VGG-16 net requires inputs of a fixed $224\times 224$ pixel size). Although there are many possible transformations of our arbitrary-shaped regions, we warp all pixels in a  bounding box around it to the required size, regardless its original size or shape. Prior to warping, we expand the tight bounding box by a certain number of pixels (10 in our system) around the original box, which was proven effective in the task of using image classifier for object detection task \cite{girshick2014}. 

After the 
classification, we collect the confidence of regions with respect to the specific classes associated with the video and form a set of scored regions,
\begin{equation} 
\{\mathcal{H}_{w_{1}}, \dots, \mathcal{H}_{w_K} \} \nonumber
\end{equation}
where
\begin{equation} 
\mathcal{H}_{w_k} = \{ (r_1, s_{r_1}, c_{r_1, w_k}), \dots, (r_N, s_{r_N}, c_{r_N, w_k}) \} \nonumber
\end{equation}
with $s_{r_i}$  is the original score of proposal $r_i$ and $c_{r_i, w_k}$ is its confidence
from CNN classification with regard to keyword or class $w_k$.  Fig. \ref{fig:diagram} shows the positive detections with confidence higher than a predefined threshold ($0.01$), where
higher confidence does not necessarily correspond to good proposals. This is mainly due to the nature of image classification where the image frame is quite often much larger than the tight bounding box of the object. In the following discussion we drop the subscript of  classes, and formulate our method with regard to one single class for the sake of clarity, albeit our method works on multiple classes.

\subsubsection{Spatial Average Pooling}

After the initial discovery, a large number of region proposals are positively detected with regard to a class label, which include overlapping regions on the same objects and spurious detections. We adopt a simple weighted spatial average pooling strategy to aggregate the region-wise score, confidence as well as their spatial extent. For each proposal $r_i$, we rescore it by multiplying its score and classification confidence, which is denoted by $\tilde{s}_{r_i} = s_{r_i} \cdot c_{r_i}$. We then
generate score map $\mathcal{S}_{r_i}$ of the size of image frame, which is composited as the binary map of current region proposal multiplied by its score $\tilde{s}_{r_i}$. We 
perform an average pooling over the score maps of all the proposals to compute a confidence map,
\begin{equation} 
\label{eq:conf1}
C^t = \frac{\sum _{r_i \in \mathcal{R}^t} \mathcal{S}_{r_i}}{\sum _{r_i \in \mathcal{R}^t} \tilde{s}_{r_i}} 
\end{equation}
where $\sum _{r_i \in \mathcal{R}^t} \mathcal{S}_{r_i}$ performs element-wise operation and $\mathcal{R}^t$ represents the set of candidate proposals from frame $t$.

\begin{figure}[t!]
	\centering
	\includegraphics[width=0.95\linewidth]{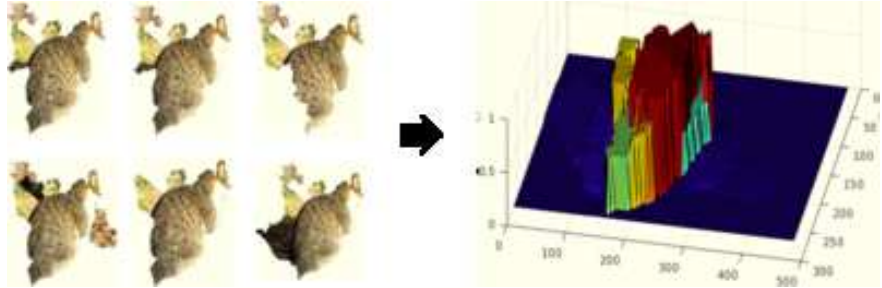}
	\caption{An illustration of the weighted spatial average pooling strategy.}\label{fig:pooling}
\end{figure}

The resulted confidence map $\mathcal{C}^t$ aggregates not only the region-wise score but also their spatial extent. The key 
insight is that good proposals coincide with each other in the spatial domain and their contribution to the final confidence map are proportional to their region-wise score. An illustration of the weighted spatial average pooling is shown in Fig. \ref{fig:pooling}.

\subsubsection{Semi-Supervised Domain Adaptation}
\label{sec:semisupervised}

To perform domain adaptation from image recognition to video object segmentation, we define a weighted space-time graph $\mathcal{G}_d=(\mathcal{V}_d,\mathcal{E}_d)$ spanning the whole video or a shot with each node corresponding to a superpixel, and each edge connecting two superpixels based on spatial and temporal adjacencies. Temporal adjacency is coarsely determined based on motion estimates, i.e., two superpixels are deemed temporally adjacent if they are connected by at least one motion vector.

We compute the affinity matrix $A$ of the graph among spatial neighbours as
\begin{equation} 
\label{eq:aff1}
A^{s} _{i,j}= \frac{ \text{exp}(-d^{c}(s_i,s_j))}{d^{s}(s_i,s_j)} 
\end{equation}
where the functions $d^{s}(s_i,s_j)$ and $d^{c}(s_i,s_j)$ computes the spatial and color distances between spatially neighbouring superpixels $s_i$ and $s_j$ respectively:
\begin{equation} 
\label{eq:colourdist}
d^{c}(s_i,s_j) = \frac{||c_i-c_j||^2}{2<||c_i-c_j||^2>} \nonumber
\end{equation}
where $||c_i-c_j||^2$ is the squared Euclidean distance between two adjacent superpixels in RGB colour space, and $<\cdot>$ computes the average over all pairs $i$ and $j$.

\begin{figure}[t!]
	\centering
	\includegraphics[width=0.95\linewidth]{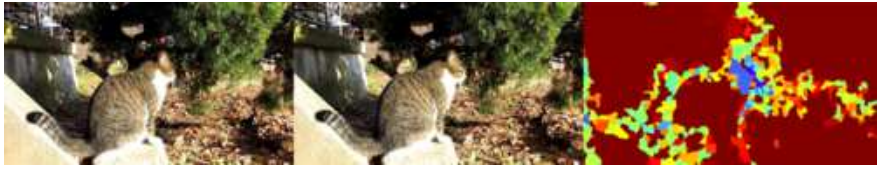}
	\caption{Motion reliability map (right) computed given the optical flow between two consecutive frames (left and middle).}\label{fig:motacc}
\end{figure}

For affinities among temporal neighbours $s_i^{t-1}$ and $s_j^t$, we consider both the temporal and colour distances between $s_i^{t-1}$ and $s_j^t$,
\begin{equation} 
\label{eq:aff2}
A^{t} _{i,j} = \frac{ \text{exp}(-d^{c}(s_i,s_j))}{d^{t}(s_i,s_j)} \nonumber
\end{equation}
where 
\begin{align} 
\label{eq:temporalcorr} 
d^{t}(s_i,s_j) &=  \frac{\rho_{i,j}}{m_{i}},\\\nonumber
m_{i} &= \mathrm{exp}(-w_c \cdot \pi_{i}), \\\nonumber
\rho_{i,j} &= \frac{|\tilde{s}_{i}^{t-1} \cap s_j^{t}|}{|\tilde{s}_{i}^{t-1}|}.\nonumber
\end{align}
Specifically, we define the temporal distance $d^{t}(s_i,s_j)$ by combining two factors, i.e., the temporal overlapping ratio $\rho_{i,j}$ and motion accuracy $m_{i}$. $\pi_i$ denotes the motion non-coherence, and $w_c=2.0$ is a parameter. The larger the temporal  overlapping ratio is between two temporally related superpixels, the closer they are in temporal domain, subject to the accuracy of motion estimation. 
The temporal overlapping ratio $\rho_{i,j}$  is defined between the warped version of $s_{i}^{t-1}$ following motion vectors and $s_j^{t}$, 
where $\tilde{s}_{i}^{t-1}$ is the warped region of $s_{i}^{t-1}$ by optical flow to frame  $t$, and $|\cdot|$ is the cardinality of a superpixel. The reliability of motion estimation inside $s_{i}^{t-1}$ is measured by the motion non-coherence. A superpixel, i.e., a small portion of a moving object, normally exhibits coherent motions. We correlate the reliability of motion estimation of a superpixel with its local motion non-coherence. We compute quantised optical flow histograms $h_{i}$ for superpixel $s_{i}^{t-1}$, and compute $\pi_i$ as the information entropy of $h_{i}$. Larger $\pi_i$ indicates higer levels of motion non-coherence, i.e., lower motion reliability of motion estimation.  An example of computed motion reliability map is shown in Fig. \ref{fig:motacc}.

We follow a similar formulation with \cite{Zhou2004} to minimise an energy function $E(X)$ with respect to all superpixels confidence $X$ ($X\in [-1, 1]$):
\begin{equation} 
\label{eq:energy}
E(X) = \sum_{i,j=1}^N A_{ij}||x_i d_{i}^{-\frac{1}{2}}-x_j d_{j}^{-\frac{1}{2}}||^2 + \mu \sum_{i=1}^N  ||x_i-c_i||^2,  
\end{equation}
where $\mu$ is the regularization parameter, and $X$ are the desirable confidence of superpixels which are  imposed by noisy confidence $C$ in Eq. (\ref{eq:conf1}).  We set $\mu=0.5$. Let the node degree matrix 
$D = \mathrm{diag}([d_1, \dots, d_N])$ be defined as $d_i=\sum_{j=1}^{N} A_{ij}$, where $N=|\mathcal{V}|$. Denoting $S = D^{-1/2}AD^{-1/2}$, 
this energy function can be minimised iteratively as $X^{t+1} = \alpha S X^t + (1-\alpha) C$ until convergence, where $\alpha$ controls the relative amount of the confidence from its neighbours and its initial confidence. Specifically, the affinity matrix $A$ of $\mathcal{G}_d$ is symmetrically normalized in $S$, which is necessary for the convergence of the following iteration. In each iteration, each superpixel adapts itself by receiving the confidence from its neighbours while preserving its initial confidence. The confidence is adapted symmetrically since $S$ is symmetric. After convergence, the confidence of each unlabeled superpixel is adapted to be the class of which it has received most confidence during the iterations.

We alternatively solve the optimization problem as a linear system of equations which is more efficient. Differentiating $E(X)$ with respect to $X$ we have
\begin{equation} 
\nabla E(X) |_{X=X^{*}} = X^{*} - SX^{*} + \mu (X^{*}-C) = 0
\end{equation}
which can be transformed as
	\begin{equation} 
	 (I - (1- \frac{\mu}{1+\mu}) S)X^{*} = \frac{\mu}{1+\mu} C.
	\end{equation}
Finally we  have 
	 \begin{equation} 
	 (I - (1-\eta) S)X^{*} = \eta C.
	 \end{equation}
where $\eta = \frac{\mu}{1+\mu}$.

The optimal solution for $X$ can be found using the preconditioned (Incomplete Cholesky factorization) conjugate gradient method with very fast convergence. For consistency, still let $C$ denote the optimal semantic confidence $X$ for the rest of this paper.

\begin{figure}[t!]
	\centering
	\subfigure[Confidence maps of three consecutive frames]{
		\includegraphics[width=0.95\linewidth]{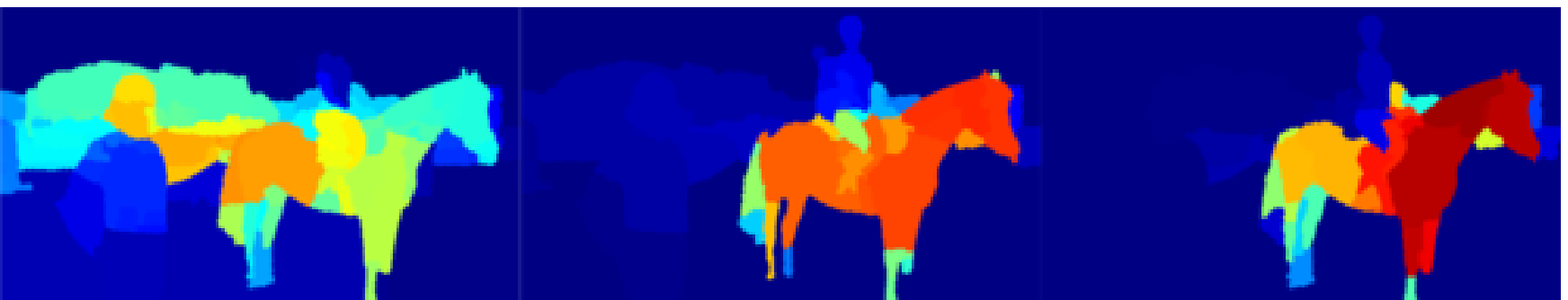}  
	}
	\subfigure[Confidence maps after domain adaptation]{
		\includegraphics[width=0.95\linewidth]{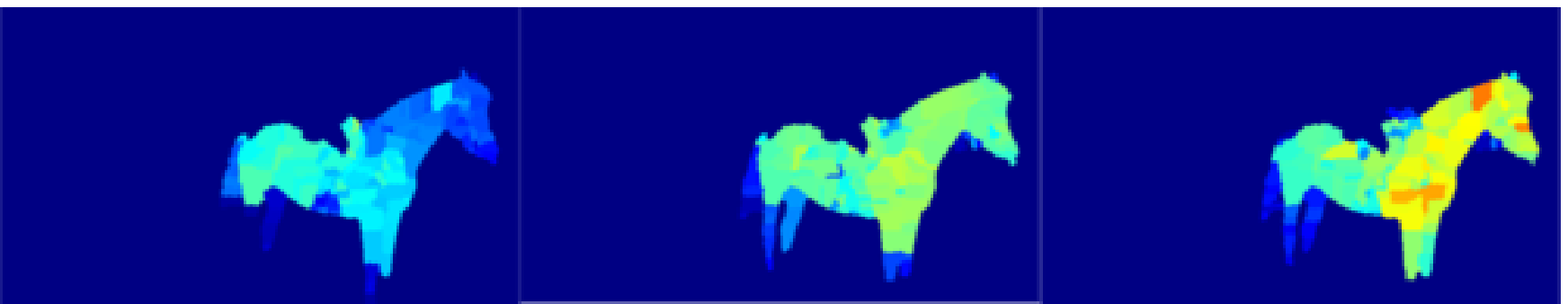}  
	}
	\caption{Proposed domain adaptation effectively adapts the noisy confidence map from image recognition to the video object segmentation domain.}
\end{figure}

\subsection{Video Object Segmentation}
\label{sec:segmentation}

We formulate video object segmentation as a superpixel-labelling problem of assigning each superpixel two classes: objects and background (not listed in the keywords). Similar to sub-sec. \ref{sec:semisupervised} we define a space-time superpixel
graph $\mathcal{G}_s=(\mathcal{V}_s,\mathcal{E}_s)$ by connecting frames temporally with optical flow displacement.

We define the energy function that minimises to achieve the optimal labeling:
\begin{equation} 
\label{eq:graphcut}
E(x) = \sum_{i\in \mathcal{V}} (\psi _{i}^{c}(x_i) + \lambda_{o} \psi _{i}^{o}(x_i)) + \lambda_{s} \sum_{i\in \mathcal{V}, j\in N_{i}^{s}} \psi _{i,j}^{s}(x_i,x_j) + \lambda_{t} \sum_{i\in \mathcal{V}, j\in N_{i}^{t}} \psi _{i,j}^{t} (x_i,x_j)
\end{equation}
where $N_{i}^{s}$ and $N_{i}^{t}$ are the sets of superpixels adjacent to superpixel $s_i$ spatially and temporally in the graph respectively; $\lambda_{o}$, $\lambda_{s}$ and $\lambda_{t}$ are parameters; $\psi _{i}^{c}(x_i)$ indicates the 
color based unary potential and $\psi _{i}^{o}(x_i)$ is the  unary potential of
semantic object confidence which measures how likely the superpixel to be labelled by $x_i$ given the semantic confidence map;  $\psi _{i,j}^{s}(x_i,x_j)$ and  $\psi _{i,j}^{t} (x_i,x_j)$ are 
spatial pairwise potential and temporal pairwise potential respectively. We set parameters $\lambda_{o} = 10$, $\lambda_{s} = 1000$ and $\lambda_{t}=2000$. 
The definitions of these unary and pairwise terms are explained in detail next.

\subsubsection{Unary Potentials}

We define unary terms to measure how likely a superpixel is to be label as background or the object of interest according to both the appearance model and semantic object confidence map. 

Colour unary potential is defined similar to \cite{RotherKB04}, which evaluates the fit of a colour distribution (of a label) to the colour of a superpixel,
\begin{equation} 
\psi _{i}^{c}(x_i) =  - \text{log} U_{i}^{c}(x_i)\nonumber
\end{equation}
where $U_{i}^{c}(\cdot)$ is the colour likelihood from colour model. 

We train two Gaussian Mixture Models (GMMs) over the RGB values of superpixels, for objects and background respectively. These GMMs are estimated by sampling the superpixel colours according to the semantic confidence map.

Semantic unary potential is defined to evaluate how likely the superpixel to be labelled by $x_i$ given the semantic confidence map $c_i^t$
\begin{equation} 
\psi _{i}^{o}(x_i) =  - \text{log} U_{i}^{o}(x_i)\nonumber
\end{equation}
where $U_{i}^{o}(\cdot)$ is the semantic likelihood, i.e., for an object labelling $U_{i}^{o} = c_i^t$ and $1-c_i^t$ otherwise.

\subsubsection{Pairwise Potentials}

We define the pairwise potentials to encourage both spatial and temporal smoothness of labelling while preserving discontinuity in the data. These terms are defined similar to the affinity matrix in sub-sec. \ref{sec:semisupervised}.

Superpixels in the same frame are spatially connected if they are adjacent. The spatial pairwise potential $\psi^{s} _{i,j}(x_i,x_j)$ penalises different labels assigned to spatially adjacent superpixels:
\begin{equation} 
\psi^{s} _{i,j}(x_i,x_j) = \frac{[x_i \neq x_j] \text{exp}(-d^{c}(s_i,s_j))}{d^{s}(s_i,s_j)} \nonumber
\end{equation}
where $[\cdot]$ denotes the indicator function.

The temporal pairwise potential is defined over edges where  superpixels are temporally connected on consecutive frames. Superpixels  $s_i^{t-1}$ and $s_j^t$ are deemed as temporally connected if there is at least one pixel 
of $s_i^{t-1}$ is propagated to $s_j^t$ following the optical flow motion vectors,
\begin{equation} 
\psi^{t} _{i,j}(x_i,x_j) = \frac{[x_i \neq x_j] \text{exp}(-d^{c}(s_i,s_j))}{d^{t}(s_i,s_j)}. \nonumber
\end{equation}
Taking advantage of the similar definitions in computing affinity matrix in sub-sec. \ref{sec:semisupervised}, the pairwise potentials can be efficiently computed by reusing the affinity in Eq. (\ref{eq:aff1}) and (\ref{eq:aff2}).

\subsubsection{Optimization}

We adopt  alpha expansion  \cite{BoykovVZ01} to minimise Eq. (\ref{eq:graphcut}) and the resulting label assignment gives the semantic object segmentation of the video. 

\subsection{Implementation}
We implement our method using MATLAB and C/C++, with Caffe \cite{jia2014caffe} implementation of VGG-16 net \cite{vggnet}. We reuse the superpixels returned from \cite{EndresH10} which is produced by \cite{ArbelaezMFM09}. Large displacement optical flow algorithm \cite{BroxBPW04} is adopted to cope with strong motion in natural videos. $5$ components per GMM in RGB colour space are learned to model the colour distribution following \cite{RotherKB04}. Our domain adaptation method performs efficient learning on superpixel graph with an unoptimised MATLAB/C++ implementation, which takes around $30$ seconds over a video shot of $100$ frames. The average time on segmenting one preprocessed frame is about $3$ seconds on a commodity desktop with a Quad-Core 4.0 GHz processor,  16 GB of RAM, and GTX 980 GPU.

We set parameters by optimizing segmentation against ground truth over a sampled set of $5$ videos from publicly available \emph{Freiburg-Berkeley Motion Segmentation Dataset} dataset \cite{brox2010object}  which proved to be a versatile setting for a wide variety of videos. These parameters are fixed for the evaluation.

\section{Evaluation}
\label{sec:evaluation}

We evaluate our method on a large scale video dataset YouTube-Objects \cite{PrestLCSF12} and SegTrack \cite{TsaiFNR12}. YouTube-Objects consists of videos from $10$ object classes with pixel-level ground truth for every $10$ frames of $126$ videos provided by \cite{jain2014supervoxel}.  These videos are very challenging and completely unconstrained, with objects of similar colour to the background, fast motion, non-rigid deformations, and fast camera motion. SegTrack consists of $5$ videos with single or interacting objects presented in each video. 

\subsection{YouTube-Objects Dataset}

We measure the segmentation performance using the standard {\em intersection-over-union} (IoU) overlap as accuracy metric. 
We compare our approach with $6$ state-of-the-art automatic approaches on this dataset, including two motion driven segmentation \cite{BroxM10,Papazoglou2013}, three weakly supervised approaches \cite{PrestLCSF12,TangSY013,ZhangCLWX15}, and 
state-of-the-art object-proposal based approach \cite{LeeKG11}. Among the compared approaches, \cite{BroxM10,LeeKG11} reported their results by fitting 
a bounding box to the largest connected segment and overlapping with the ground-truth bounding box; the result of \cite{LeeKG11} on this dataset is originally reported by \cite{Papazoglou2013} by testing on $50$ videos ($5$/class). The performance of \cite{Papazoglou2013} measured with respect to segmentation groundtruth is reported by \cite{ZhangCLWX15}.  Zhang  {\em et al.} \cite{ZhangCLWX15} reported results in more than 5500 frames sampled in the dataset based on the segmentation groundtruth. Wang {\em et al.} \cite{WangSP15}  reported the average results on $12$ randomly sampled videos in terms of a different metric, i.e., per-frame pixel errors across all categories, and thus not listed here for comparison.

\begin{table}[t!]
	\caption{Intersection-over-union overlap accuracies on YouTube-Objects Dataset}
	\centering 
	\small
	\begin{tabular}{| c | c | c | c | c | c | c | c |  c |} 
		\hline
		& \shortstack{Brox\\ \cite{BroxM10} } & \shortstack{Lee\\ \cite{LeeKG11}} & \shortstack{Prest\\ \cite{PrestLCSF12}} & \shortstack{Papazoglou\\ \cite{Papazoglou2013}} & \shortstack{Tang\\ \cite{TangSY013}}  &  \shortstack{Zhang\\ \cite{ZhangCLWX15}} & Baseline & Ours\\
		\hline
		Plane & 0.539 & NA  & 0.517  & 0.674  & 0.178  & \underline{\textbf{0.758}}   & 0.693      & 0.757\\
		Bird    & 0.196 & NA  & 0.175  & 0.625  & 0.198  & 0.608  & 0.590      & \underline{\textbf{0.658}} \\
		Boat   & 0.382 & NA  & 0.344  & 0.378  & 0.225  & 0.437   & 0.564     & \underline{\textbf{0.656}} \\
		Car     & 0.378 & NA  & 0.347  & 0.670  & 0.383  &  \underline{\textbf{0.711}}    & 0.594     & 0.650\\
		Cat	  & 0.322 & NA  & 0.223  & 0.435  & 0.236  & 0.465    & 0.455    &  \underline{\textbf{0.514}} \\
		Cow   & 0.218 & NA  & 0.179  & 0.327   & 0.268  & 0.546   & 0.647      &  \underline{\textbf{0.714}} \\
		Dog    & 0.270 & NA  & 0.135  & 0.489   & 0.237  & 0.555  & 0.495      & \underline{\textbf{0.570}} \\
		Horse & 0.347 & NA  & 0.267  & 0.313   & 0.140  & 0.549   & 0.486      &  \underline{\textbf{0.567}} \\
		Mbike & 0.454 & NA  & 0.412  & 0.331   & 0.125  & 0.424   & 0.480  	&  \underline{\textbf{0.560}} \\
		Train   & 0.375 & NA & 0.250  & \underline{\textbf{0.434}}    & 0.404  & 0.358	& 0.353 	&  0.392 \\
		\hline
		Cls. Avg. &0.348 &0.28 &0.285 &0.468   & 0.239  & 0.541  & 0.536    	& \underline{\textbf{0.604}} \\
		\hline
		Vid. Avg. & NA & NA & NA &0.432   &  0.228  & 0.526   & 0.523   	& \underline{\textbf{0.592}} \\
		\hline
	\end{tabular} \label{tbl:yto-result} 
\end{table}

As shown in Table \ref{tbl:yto-result}, our method  outperforms the competing methods in  $7$ out of $10$ classes, with gains up to $6.3\%$/$6.6\%$ in category/video average accuracy over the best competing method \cite{ZhangCLWX15}. This is remarkable considering that \cite{ZhangCLWX15} employed strongly-supervised deformable part models (DPM) as object detector while our approach only leverages image recognition model which lacks the capability of localizing objects. \cite{ZhangCLWX15}  outperforms our method on \emph{Plane} and \emph{Car}, otherwise exhibiting varying performence across the categories --- higher accuracy on more rigid objects but lower accuracy on highly flexible and deformable objects such as \emph{Cat} and \emph{Dog}. We owe it to that, though based on object detection, \cite{ZhangCLWX15}  prunes noisy detections and regions by enforcing spatio-temporal constraints, rather than learning an adapted data-driven representation  in our approach. It is also worth remarking on the improvement in classes, e.g.,  \emph{Cow},  where the existing methods normally fail or underperform due to the heavy reliance on motion information. The main challenge of the \emph{Cow} videos is that cows very frequently stand still or move with mild motion, which the existing approaches might fail to capture whereas our proposed method excels by leveraging the  recognition and representation power of deep convolutional neural network, as well as the semi-supervised domain adaptation. 

Interestingly, another weakly supervised method \cite{TangSY013} slightly outperforms our method on \emph{Train} although all  methods do not perform very well on this category due to the slow motion and missed detections on partial views of trains. This is probably owing to that \cite{TangSY013} uses a large number of similar training videos which may capture objects in rare view. Otherwise, our method doubles or triples the accuracy of \cite{TangSY013}. Motion driven method  \cite{Papazoglou2013} can
better distinguish rigid moving foreground objects on videos exhibiting relatively clean backgrounds, such as \emph{Plane} and \emph{Car}.

Comparing with the baseline scheme, we can see the proposed semi-supervised domain adaptation is able to learn to successfully adapt to the target with a gain of $6.8\%$/$6.9\%$ in category/video average accuracies. 

\begin{figure}[h!]
\centering
\subfigure[Aeroplane]{
	\includegraphics[width=0.17\linewidth,height=5cm]{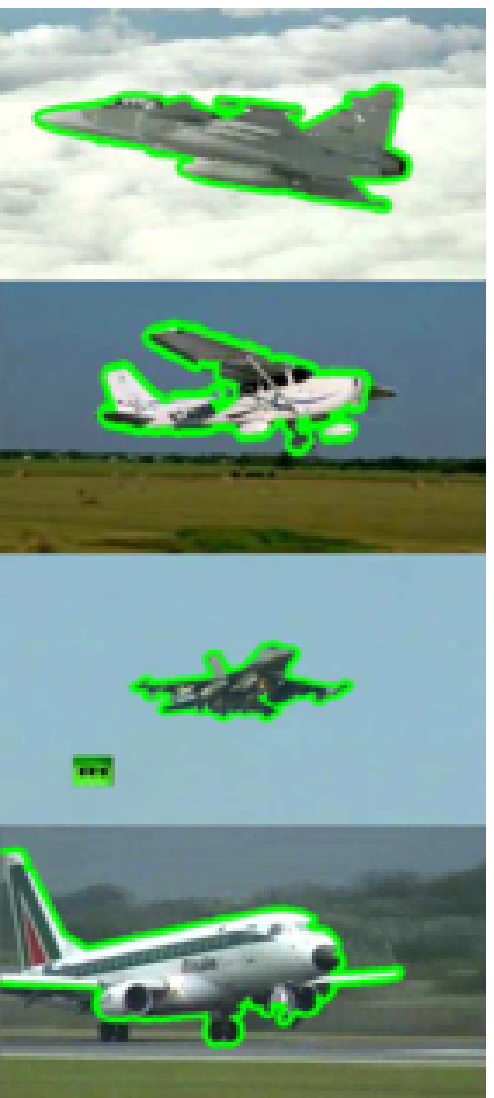}  
}
\subfigure[Bird]{
	\includegraphics[width=0.17\linewidth,height=5cm]{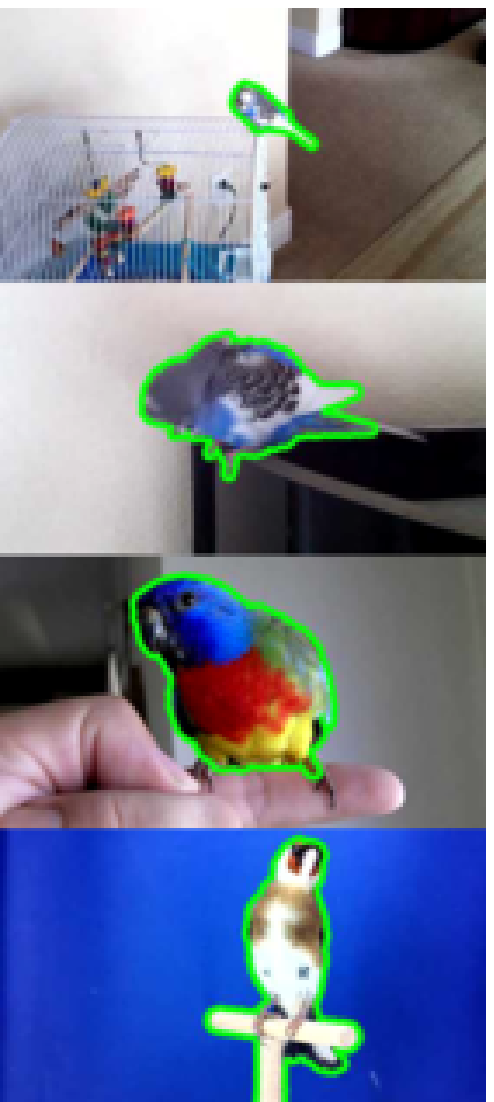}  
}
\subfigure[Boat]{
	\includegraphics[width=0.17\linewidth,height=5cm]{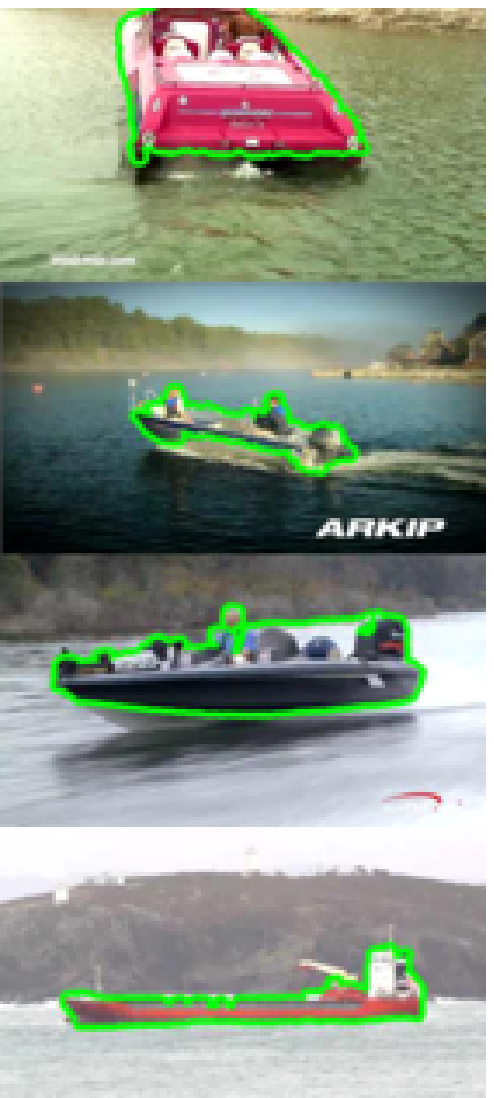}  
}
\subfigure[Car]{
	\includegraphics[width=0.17\linewidth,height=5cm]{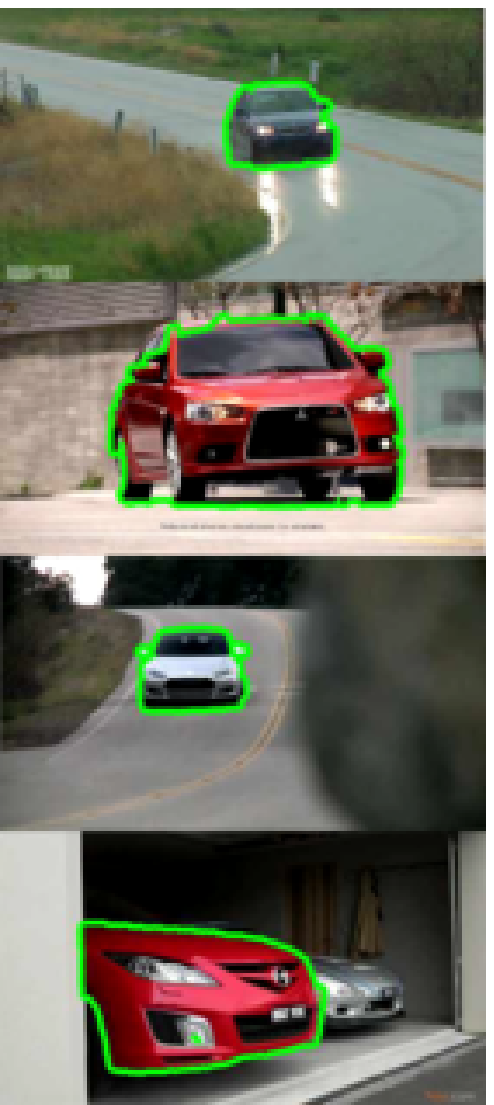}  
}
\subfigure[Cat]{
	\includegraphics[width=0.17\linewidth,height=5cm]{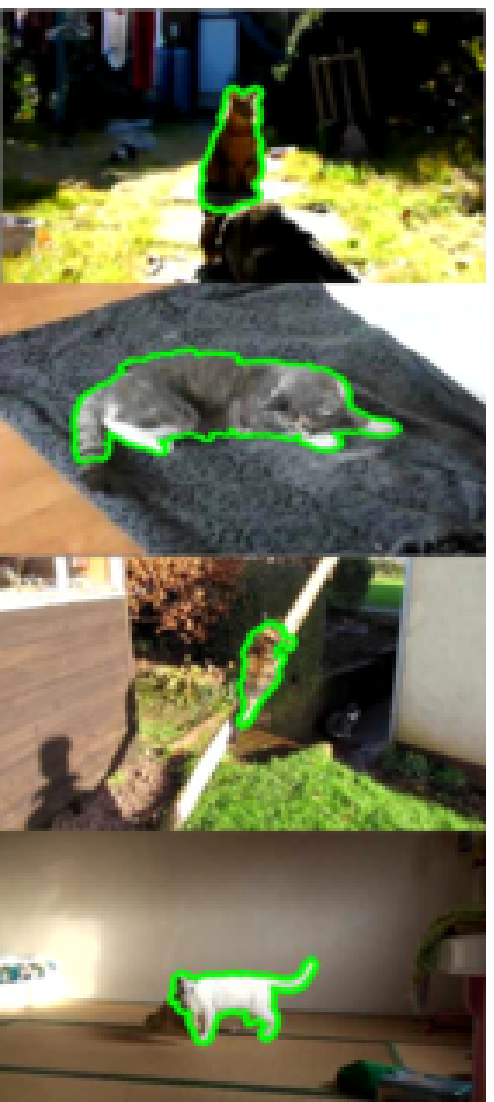}  
}
\subfigure[Cow]{
	\includegraphics[width=0.17\linewidth,height=5cm]{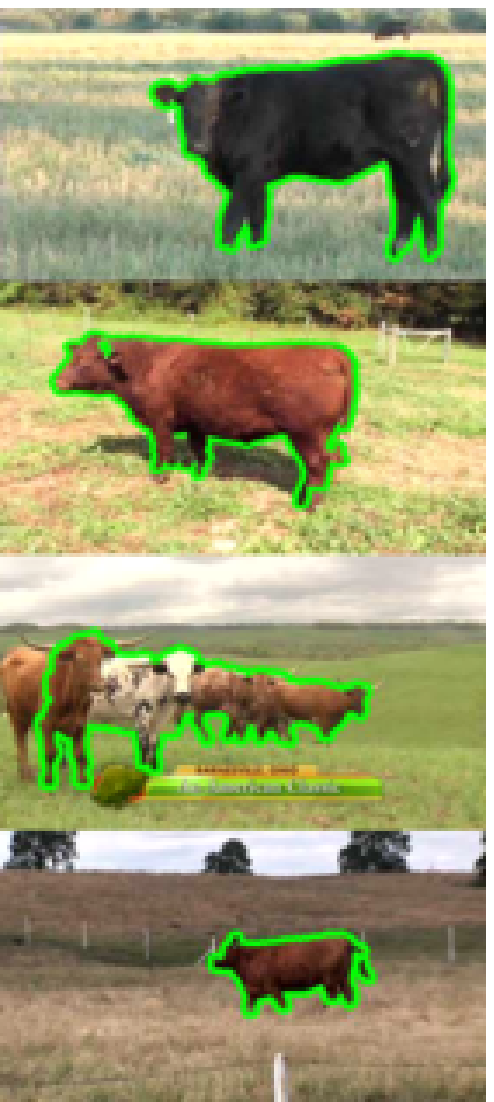}  
}
\subfigure[Dog]{
	\includegraphics[width=0.17\linewidth,height=5cm]{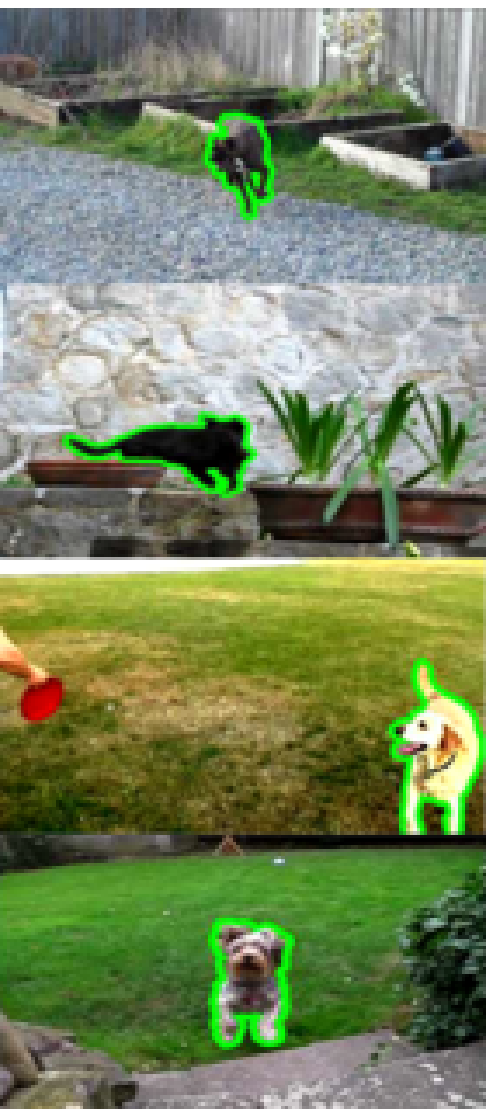}  
}
\subfigure[Horse]{
	\includegraphics[width=0.17\linewidth,height=5cm]{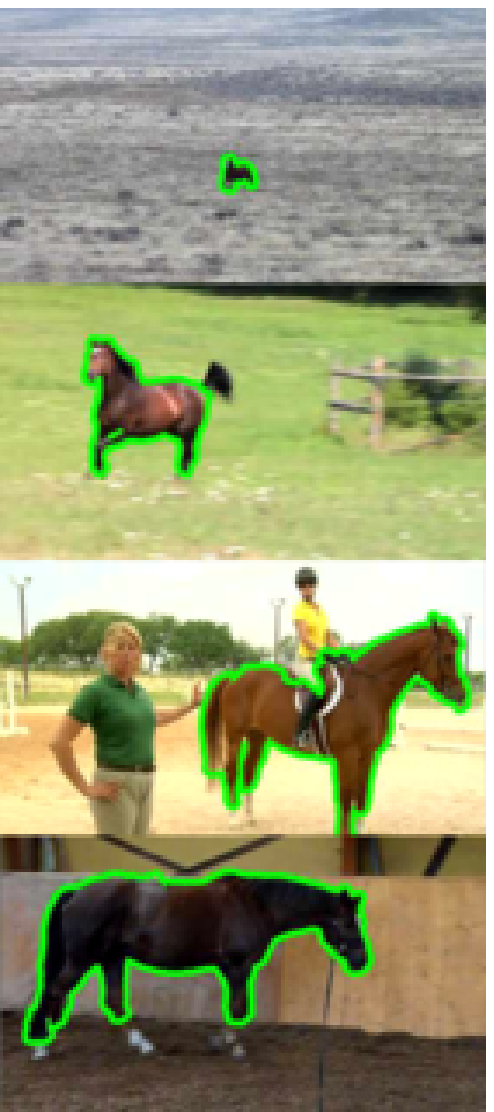}  
}
\subfigure[Motorbike]{
	\includegraphics[width=0.17\linewidth,height=5cm]{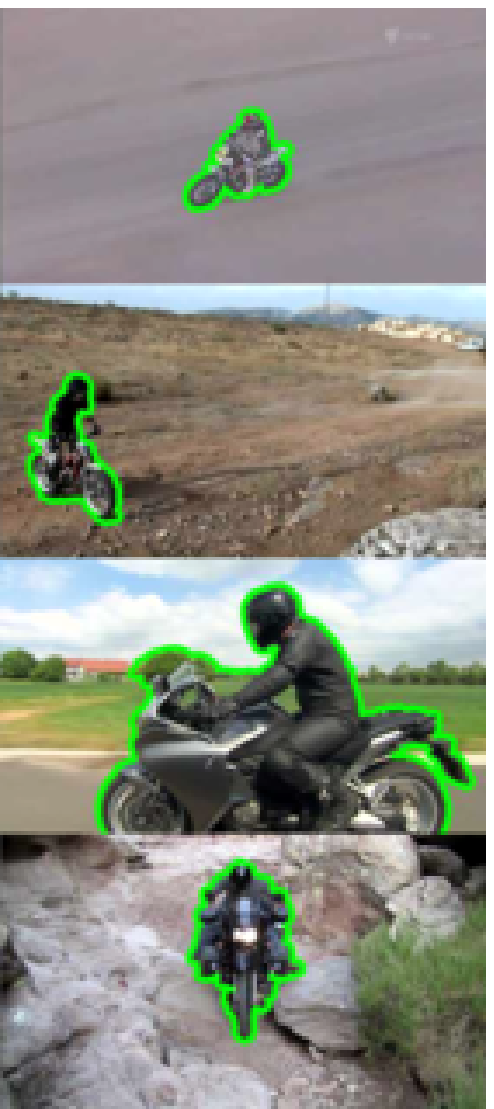}  
}
\subfigure[Train]{
	\includegraphics[width=0.17\linewidth,height=5cm]{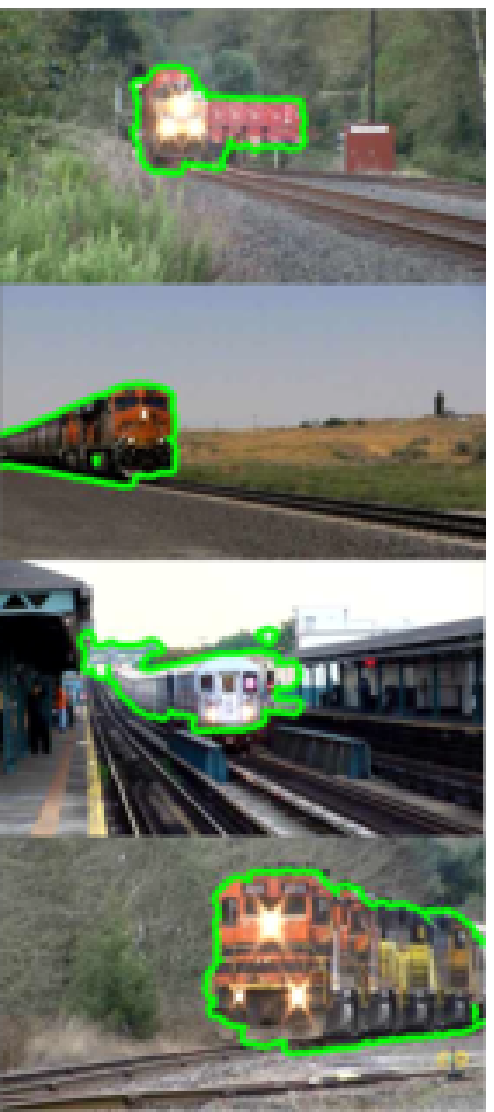}  
}
\caption{Representative successful results by our approach on YouTube-Objects dataset.}
\end{figure}

\subsection{SegTrack Dataset}
 
 We evaluate on SegTrack dataset to compare with the representative state-of-the-art unsupervised object segmentation algorithms  \cite{BroxM10,LeeKG11,ZhangJS13,Papazoglou2013,ZhangCLWX15}. To avoid confusion of segmentation results, all the compared methods only consider the primary object.

\begin{figure}[t!]
	\centering
	\includegraphics[width=0.95\linewidth]{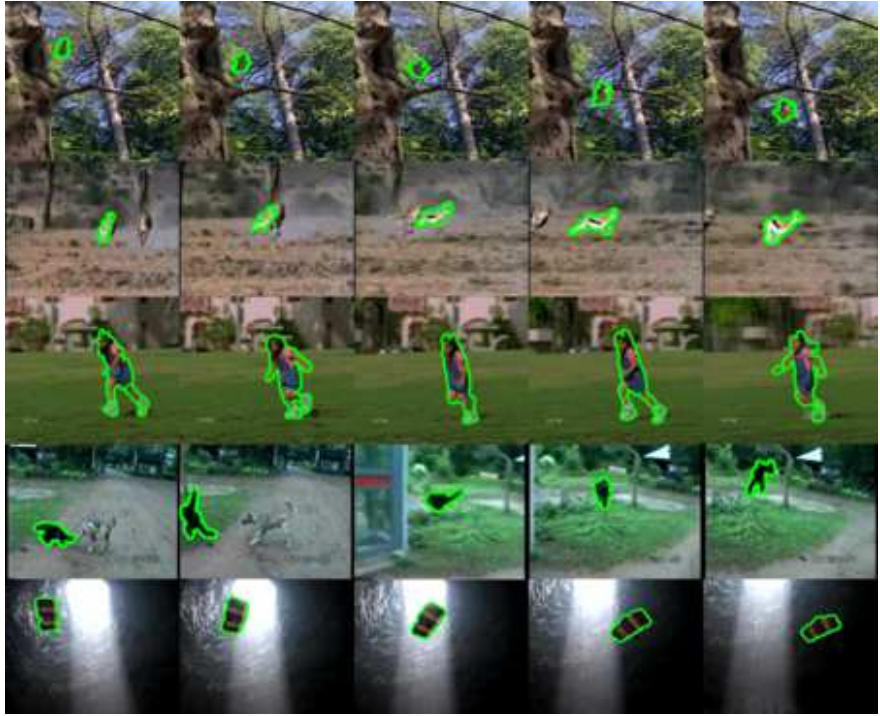}
	\caption{Qualitative results of our method on SegTrack dataset.}\label{fig:segtrack}
\end{figure}

As shown in Table \ref{tbl:segtrack}, our method outperforms weakly supervised method  \cite{ZhangCLWX15} on \emph{birdfall} and \emph{monkeydog} videos,  motion driven method \cite{Papazoglou2013} on four out of five videos, and proposal ranking method  \cite{LeeKG11} on four videos. Clustering point tracks based method \cite{BroxM10} results in highest error among all the  methods. Overall, our performance is about on par with weakly supervised method  \cite{ZhangCLWX15}. The proposal merging method \cite{ZhangJS13} obtains best results on three videos, yet it is sensitive to motion accuracy as reported by \cite{WangW14} on other dataset.  We believe that the progress on this dataset is plateaued due to the limited number of available video sequences. Qualitative segmentation of our approach is shown in Fig. \ref{fig:segtrack}.

\begin{table}[t!]
	\caption{Quantitative segmentation results on SegTrack. Segmentation error as measured by the average number of incorrect pixels per frame. } 
	\centering 
	\begin{tabular}{| c | c | c | c | c | c | c |  c |} 
		\hline
		Video (No. frames) & Ours & \cite{BroxM10} & \cite{Papazoglou2013} & \cite{ZhangJS13}  & \cite{LeeKG11} & \cite{ZhangCLWX15}  \\
		\hline
		birdfall (30) & 170 &  468 & 217  & 155 &  288 & 339\\
		cheetah (29) & 826  & 1968 & 890  & 633 &  905 & 803\\
		girl (21) & 1647 & 7595 & 3859  & 1488 & 1785 & 1459\\
		monkeydog (71) & 304  & 1434& 284  &  472 & 521 & 365\\
		parachute (51) & 363 & 1113 & 855  & 220 &  201 & 196\\	
		\hline
	\end{tabular} \label{tbl:segtrack} 
\end{table}

\section{Conclusion}
\label{sec:conclusion}
We have proposed a semi-supervised framework to adapt CNN classifiers from image recognition domain to the target domain of semantic video object segmentation. This framework combines the recognition and representation power of CNN with the intrinsic structure of unlabelled data in the target domain to improve inference performance, imposing spatio-temporal smoothness constraints on the semantic confidence over the unlabeled video data. This proposed domain adaptation framework enables learning a data-driven representation of video objects. We demonstrated that this representation underpins a robust semantic video object segmentation method which outperforms existing methods on challenging datasets. As a future work, it would be interesting to incorporate representations learned from higher layers of CNN into the domain adaptation, which might potentially improve adaptation by propagating and combining higher level context.

\bibliographystyle{splncs}
\bibliography{refs}

\begin{thebibliography}{10}

\bibitem{BroxM10}
Brox, T., Malik, J.:
\newblock Object segmentation by long term analysis of point trajectories.
\newblock In: ECCV. (2010)  282--295

\bibitem{LeeKG11}
Lee, Y.J., Kim, J., Grauman, K.:
\newblock Key-segments for video object segmentation.
\newblock In: ICCV. (2011)  1995--2002

\bibitem{ZhangJS13}
Zhang, D., Javed, O., Shah, M.:
\newblock Video object segmentation through spatially accurate and temporally
  dense extraction of primary object regions.
\newblock In: CVPR. (2013)  628--635

\bibitem{Papazoglou2013}
Papazoglou, A., Ferrari, V.:
\newblock Fast object segmentation in unconstrained video.
\newblock In: ICCV. (2013)  1777--1784

\bibitem{WangW14}
Wang, T., Wang, H.:
\newblock Graph transduction learning of object proposals for video object
  segmentation.
\newblock In: ACCV. (2014)  553--568

\bibitem{wang2016primary}
Wang, H., Wang, T.:
\newblock Primary object discovery and segmentation in videos via graph-based
  transductive inference.
\newblock Computer Vision and Image Understanding \textbf{143} (2016)  159--172

\bibitem{alexnet}
Krizhevsky, A., Sutskever, I., Hinton, G.E.:
\newblock Imagenet classification with deep convolutional neural networks.
\newblock In: NIPS. (2012)  1106--1114

\bibitem{vggnet}
Simonyan, K., Zisserman, A.:
\newblock Very deep convolutional networks for large-scale image recognition.
\newblock arXiv preprint arXiv:1409.1556 (2014)

\bibitem{laddernet}
Rasmus, A., Valpola, H., Honkala, M., Berglund, M., Raiko, T.:
\newblock Semi-supervised learning with ladder network.
\newblock In: NIPS. (2015)

\bibitem{Yang2015}
Yang, J., Zhao, G., Yuan, J., Shen, X., Lin, Z., Price, B., Brandt, J.:
\newblock Discovering primary objects in videos by saliency fusion and
  iterative appearance estimation.
\newblock {IEEE} Trans. Circuits Syst. Video Technol. (2015)

\bibitem{EndresH10}
Endres, I., Hoiem, D.:
\newblock Category independent object proposals.
\newblock In: ECCV. (2010)  575--588

\bibitem{ManenGG13}
Manen, S., Guillaumin, M., Gool, L.J.V.:
\newblock Prime object proposals with randomized prim's algorithm.
\newblock In: ICCV. (2013)  2536--2543

\bibitem{WangXSC04}
Wang, J., Xu, Y., Shum, H.Y., Cohen, M.F.:
\newblock Video tooning.
\newblock ACM Trans. Graph. \textbf{23} (2004)  574--583

\bibitem{CollomosseRH05}
Collomosse, J.P., Rowntree, D., Hall, P.M.:
\newblock Stroke surfaces: Temporally coherent artistic animations from video.
\newblock IEEE Trans. Vis. Comput. Graph. \textbf{11} (2005)  540--549

\bibitem{WangC12}
Wang, T., Collomosse, J.P.:
\newblock Probabilistic motion diffusion of labeling priors for coherent video
  segmentation.
\newblock IEEE Transactions on Multimedia \textbf{14} (2012)  389--400

\bibitem{TsaiFNR12}
Tsai, D., Flagg, M., Nakazawa, A., Rehg, J.M.:
\newblock Motion coherent tracking using multi-label mrf optimization.
\newblock International Journal of Computer Vision \textbf{100} (2012)
  190--202

\bibitem{Li2013}
Li, F., Kim, T., Humayun, A., Tsai, D., Rehg, J.M.:
\newblock Video segmentation by tracking many figure-ground segments.
\newblock In: ICCV, Australia, December 1-8, 2013. (2013)  2192--2199

\bibitem{WangHC14}
Wang, T., Han, B., Collomosse, J.P.:
\newblock Touchcut: Fast image and video segmentation using single-touch
  interaction.
\newblock Computer Vision and Image Understanding \textbf{120} (2014)  14--30

\bibitem{GrundmannKHE10a}
Grundmann, M., Kwatra, V., Han, M., Essa, I.A.:
\newblock Efficient hierarchical graph-based video segmentation.
\newblock In: CVPR. (2010)  2141--2148

\bibitem{XuXC12}
Xu, C., Xiong, C., Corso, J.J.:
\newblock Streaming hierarchical video segmentation.
\newblock In: ECCV (6). (2012)  626--639

\bibitem{WangGP09}
Wang, C., de~La~Gorce, M., Paragios, N.:
\newblock Segmentation, ordering and multi-object tracking using graphical
  models.
\newblock In: ICCV. (2009)  747--754

\bibitem{SundbergBMAM11}
Sundberg, P., Brox, T., Maire, M., Arbelaez, P., Malik, J.:
\newblock Occlusion boundary detection and figure/ground assignment from
  optical flow.
\newblock In: CVPR. (2011)  2233--2240

\bibitem{GiordanoMPS15}
Giordano, D., Murabito, F., Palazzo, S., Spampinato, C.:
\newblock Superpixel-based video object segmentation using perceptual
  organization and location prior.
\newblock In: CVPR. (2015)  4814--4822

\bibitem{TaylorKS15}
Taylor, B., Karasev, V., Soatto, S.:
\newblock Causal video object segmentation from persistence of occlusions.
\newblock In: CVPR. (2015)  4268--4276

\bibitem{WangSP15}
Wang, W., Shen, J., Porikli, F.:
\newblock Saliency-aware geodesic video object segmentation.
\newblock In: CVPR. (2015)  3395--3402

\bibitem{HartmannGHTKMVERS12}
Hartmann, G., Grundmann, M., Hoffman, J., Tsai, D., Kwatra, V., Madani, O.,
  Vijayanarasimhan, S., Essa, I.A., Rehg, J.M., Sukthankar, R.:
\newblock Weakly supervised learning of object segmentations from web-scale
  video.
\newblock In: ECCV Workshop. (2012)  198--208

\bibitem{TangSY013}
Tang, K.D., Sukthankar, R., Yagnik, J., Li, F.:
\newblock Discriminative segment annotation in weakly labeled video.
\newblock In: CVPR. (2013)  2483--2490

\bibitem{LiuTSRCB14}
Liu, X., Tao, D., Song, M., Ruan, Y., Chen, C., Bu, J.:
\newblock Weakly supervised multiclass video segmentation.
\newblock In: CVPR. (2014)  57--64

\bibitem{ZhangCLWX15}
Zhang, Y., Chen, X., Li, J., Wang, C., Xia, C.:
\newblock Semantic object segmentation via detection in weakly labeled video.
\newblock In: CVPR. (2015)  3641--3649

\bibitem{girshick2014}
Girshick, R., Donahue, J., Darrell, T., Malik, J.:
\newblock Rich feature hierarchies for accurate object detection and semantic
  segmentation.
\newblock In: CVPR. (2014)  580--587

\bibitem{Zhou2004}
Zhou, D., Bousquet, O., Lal, T.N., Weston, J., Sch, B.:
\newblock Learning with local and global consistency.
\newblock In: NIPS. (2004)  321--328

\bibitem{RotherKB04}
Rother, C., Kolmogorov, V., Blake, A.:
\newblock "grabcut": interactive foreground extraction using iterated graph
  cuts.
\newblock ACM Trans. Graph. \textbf{23} (2004)  309--314

\bibitem{BoykovVZ01}
Boykov, Y., Veksler, O., Zabih, R.:
\newblock Fast approximate energy minimization via graph cuts.
\newblock IEEE Trans. Pattern Anal. Mach. Intell. \textbf{23} (2001)
  1222--1239

\bibitem{jia2014caffe}
Jia, Y., Shelhamer, E., Donahue, J., Karayev, S., Long, J., Girshick, R.,
  Guadarrama, S., Darrell, T.:
\newblock Caffe: Convolutional architecture for fast feature embedding.
\newblock In: Proceedings of the ACM International Conference on Multimedia,
  ACM (2014)  675--678

\bibitem{ArbelaezMFM09}
Arbelaez, P., Maire, M., Fowlkes, C.C., Malik, J.:
\newblock From contours to regions: An empirical evaluation.
\newblock In: CVPR. (2009)  2294--2301

\bibitem{BroxBPW04}
Brox, T., Bruhn, A., Papenberg, N., Weickert, J.:
\newblock High accuracy optical flow estimation based on a theory for warping.
\newblock In: ECCV. (2004)  25--36

\bibitem{brox2010object}
Brox, T., Malik, J.:
\newblock Object segmentation by long term analysis of point trajectories.
\newblock In: ECCV.
\newblock Springer (2010)  282--295

\bibitem{PrestLCSF12}
Prest, A., Leistner, C., Civera, J., Schmid, C., Ferrari, V.:
\newblock Learning object class detectors from weakly annotated video.
\newblock In: CVPR. (2012)  3282--3289

\bibitem{jain2014supervoxel}
Jain, S.D., Grauman, K.:
\newblock Supervoxel-consistent foreground propagation in video.
\newblock In: ECCV.
\newblock Springer (2014)  656--671

\end{thebibliography}

\end{document}